\documentclass[letterpaper, 10 pt, conference]{ieeeconf}  

\IEEEoverridecommandlockouts                              

\usepackage{graphics} 
\usepackage{epsfig} 
\usepackage{mathptmx} 
\usepackage{times} 
\usepackage{amsmath} 
\usepackage{amssymb}  

\usepackage{subcaption}
\usepackage{caption}
\usepackage{tabu}                     
\usepackage{multicol}                 
\usepackage{multirow}                
\usepackage{float}                    
\usepackage{makecell}                 
\usepackage{booktabs}                 
\title{\LARGE \bf
CasPoinTr: Point Cloud Completion with Cascaded Networks and Knowledge Distillation
}

\author{Yifan Yang, Yuxiang Yan, Boda Liu, Jian Pu$^{\ast}$
\thanks{All authors are with Institute of Science and Technology for Brain-Inspired Intelligence (ISTBI), Fudan University, Shanghai 200433, China (E-mails: \{yifyang23, yxyan22, bdliu22 \}@m.fudan.edu.cn, jianpu@fudan.edu.cn).
        }%
\thanks{Corresponding author: Jian Pu.}%
}

\begin{document}

\maketitle
\thispagestyle{empty}
\pagestyle{empty}

\begin{abstract}

Point clouds collected from real-world environments are often incomplete due to factors such as limited sensor resolution, single viewpoints, occlusions, and noise. These challenges make point cloud completion essential for various applications. A key difficulty in this task is predicting the overall shape and reconstructing missing regions from highly incomplete point clouds. To address this, we introduce CasPoinTr, a novel point cloud completion framework using cascaded networks and knowledge distillation. CasPoinTr decomposes the completion task into two synergistic stages: Shape Reconstruction, which generates auxiliary information, and Fused Completion, which leverages this information alongside knowledge distillation to generate the final output. Through knowledge distillation, a teacher model trained on denser point clouds transfers incomplete-complete associative knowledge to the student model, enhancing its ability to estimate the overall shape and predict missing regions. Together, the cascaded networks and knowledge distillation enhance the model’s ability to capture global shape context while refining local details, effectively bridging the gap between incomplete inputs and complete targets. Experiments on ShapeNet-55 under different difficulty settings demonstrate that CasPoinTr outperforms existing methods in shape recovery and detail preservation, highlighting the effectiveness of our cascaded structure and distillation strategy.

\end{abstract}

\section{INTRODUCTION}
Point clouds effectively capture the three-dimensional geometry and surface details of objects or scenes through their dense spatial distribution. However, in real-world scenarios, point clouds are frequently incomplete due to factors such as sensor occlusions, environmental disturbances, and hardware constraints. To overcome these challenges, point cloud completion techniques\cite{p2c, SSC} have been developed to reconstruct the complete geometric and topological structures of objects. \par
The goal of point cloud completion is to enable the model to learn the mapping relationship between incomplete and complete point clouds. Research achievements\cite{point3, seedformer} in point cloud completion have demonstrated that models can effectively learn such mapping relationships. Recently, PoinTr\cite{pointr} improves the model's understanding of missing regions by predicting center points, thereby optimizing the final completion results. However, we believe the current solution can be further refined. When addressing point cloud completion, humans typically first infer the overall shape of the complete point cloud from the incomplete input and then predict the details of the missing regions. Inspired by this cognitive process, we propose that the model should first accurately understand the shape of the complete point cloud before predicting the center points of missing regions and reconstructing the complete point cloud. Based on this, incorporating additional auxiliary information becomes a feasible approach.\par
Prototype learning\cite{proto_p_1, proto_p_3} is the method that learns representative prototypes for each class in the embedding space and measures the distance between input features and these learned prototypes. ProtoTransfer introduces a multimodal learning framework to learn class prototypes aligned in feature space, aiming to facilitate knowledge transfer across modalities\cite{proto_p_2}. Xu et al.\cite{cp3} propose a Semantic Conditional Refinement network that modulates multi-scale refinement guided by semantic information. Although these methods enhance point cloud completion by leveraging auxiliary information, they depend on additional category information and learnable prototypes rather than directly utilizing the raw point cloud data.\par

\begin{figure*}[t]
\centering
\includegraphics[width=\textwidth]{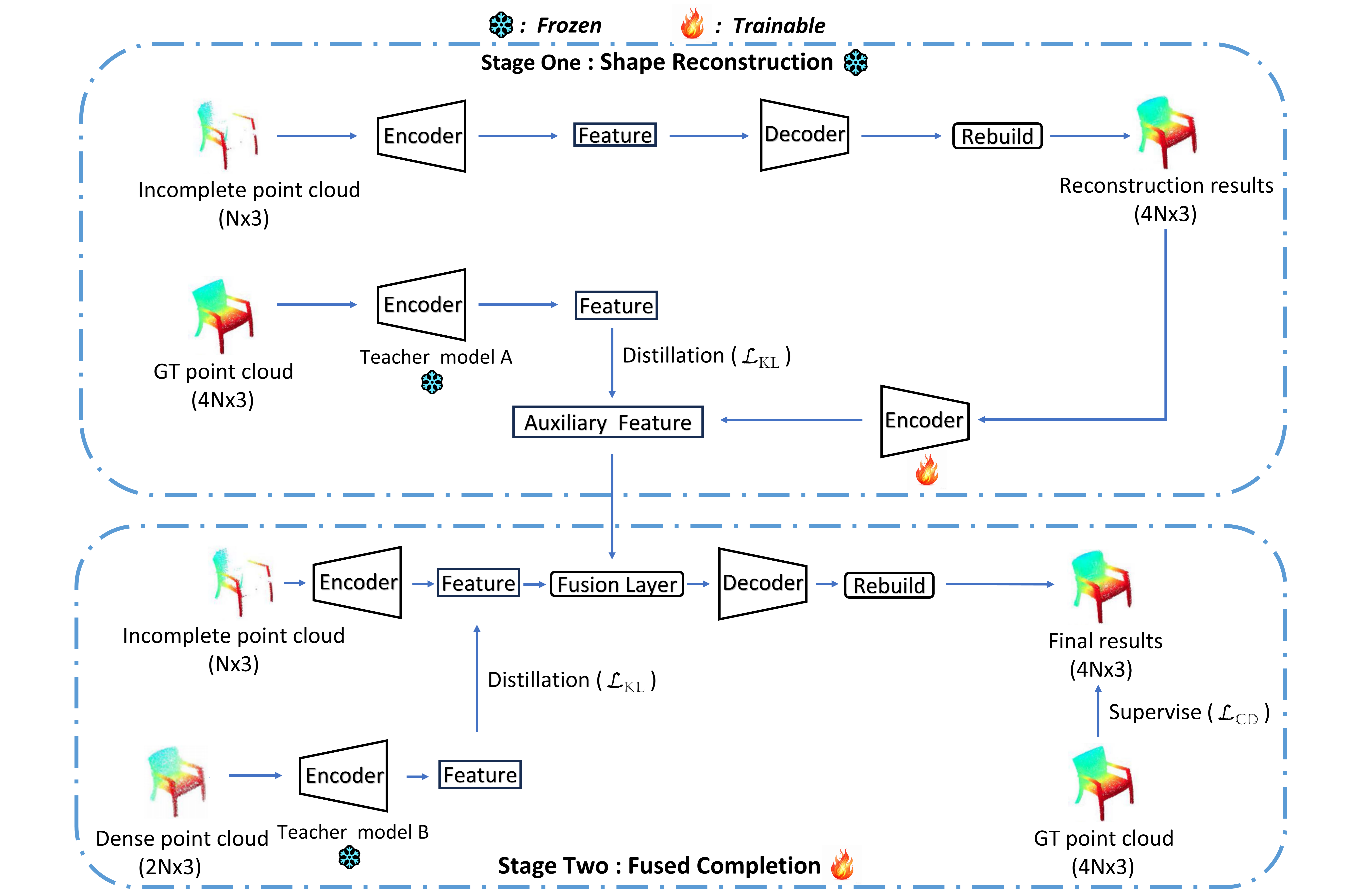}
\caption{The framework of CasPoinTr. CasPoinTr adopts AdaPoinTr as the backbone and incorporates knowledge distillation and cascaded networks. The cascaded networks split point cloud completion into two stages: Shape Reconstruction and Fused Completion. Shape Reconstruction aims to recover information of dense point clouds and provides auxiliary information to guide Fused Completion. This enables the model in the Fused Completion stage to gain a more accurate understanding of the point cloud. A teacher model with privileged inputs of high-resolution point clouds is employed for knowledge distillation, allowing the encoder to learn the incomplete-complete association and extract more informative features.}
\label{fig:caspointr}
\end{figure*}

Inspired by the successful application of cascaded networks\cite{cas_refine, pmp} and knowledge distillation\cite{dist3, dist6} in compter vison, we introduce these techniques to provide more information about the complete point cloud as guidance during encoding stage. Our goal is to enable the model to learn the mapping relationship between incomplete point clouds and latent features, ensuring that the latent features contain not only information from the input point cloud but also the target shape for completion.\par
However, existing methods face two key challenges. First, cascaded networks suffer from error accumulation. When the completion process is divided into multiple stages, errors originating in earlier stages can propagate to subsequent ones, ultimately degrading the quality of the final output. Second, in knowledge distillation, an excessive gap between the teacher-student paradigms can hinder effective knowledge transfer\cite{dist_gap}. For instance, when the teacher model uses high-resolution point cloud as privileged input while the student model only has access to low-resolution input, the student model would struggle to learn unique information from the teacher due to the increased difficulty. Additionally, inproper knowledge distillation design can cause shortcut issue\cite{shortcut}. Specifically, during the pre-training of the teacher model, directly feeding ground truth (GT) into the model may cause the model to take shortcuts in mapping inputs to outputs, reducing the accuracy and reliability of the features encoded by the teacher model. \par
To overcome these challenges, we propose CasPoinTr, a novel approach that integrates auxiliary completion and carefully designed privileged inputs for the teacher model, as depicted in Fig.\ref{fig:caspointr}. Our method divides the point cloud completion process into two stages: Shape Reconstruction and  Fused Completion. Shape Reconstruction is the initial stage that generates an intermediate representation of the point cloud. Fused Completion is the subsequent stage that utilizes the output from Shape Reconstruction as auxiliary information to guide its process, rather than directly adopting it as the point cloud input.
In addition, we employ knowledge distillation to transfer incomplete-complete associative knowledge from a teacher model, which has access to appropriately resolutioned point clouds, to the student model during training. This dual strategy enables the Fused Completion model to leverage auxiliary information from Shape Reconstruction while benefiting from knowledge distillation. As a result, the model can more effectively estimate the overall shape from incomplete point clouds and predict missing regions with greater accuracy.\par
To evaluate the effectiveness of CasPoinTr, we conducted extensive experiments on ShapeNet-55\cite{shapenet}. The results highlight the substantial improvements achieved through our cascaded networks design and knowledge distillation strategy. Our contributions can be summarized as:\\
\begin{itemize}

\item We introduce a cascaded networks structure based on auxiliary completion. It provides information about the completion target while mitigating the impact of error accumulation.
\item We design a training strategy based on knowledge distillation, which effectively addresses both the teacher-student paradigm gap and the shortcut problem without relying on excessively complex models or additional data sources. 
\item We conduct extensive experiments on ShapeNet-55 under different difficulty settings. Our study provides a thorough exploration of the detailed design of cascaded networks and knowledge distillation. The experimental results demonstrate that CasPoinTr outperforms previous state-of-the-art models.

\end{itemize}

\section{RELATED WORK}
\subsection{Point Cloud Completion}
With the advancement of deep learning, learning-based methods\cite{3D_1, 3D_2} began to utilize 3D convolutional neural networks (3D-CNNs) for point cloud completion tasks. PointNet and its variants\cite{pointnet, pointnet++} emerged as pioneering works that directly process 3D point cloud data, inspiring research in many downstream tasks. The proposal of PCN\cite{pcn} marked the beginning of a new era in point cloud completion tasks. Subsequently, many methods\cite{point1, point2} improved completion results by leveraging local features to preserve the observed geometric details from incomplete point clouds. Meanwhile, learning-based point cloud sampling methods\cite{S-NET, GS-NET} emerged and continuously evolved, enabling the preservation of point cloud features during the sampling process. Recently, the emergence of various methods\cite{grnet, snowflake} has continuously improved the results of point cloud completion. \par
More recently, Yu et al. proposed PoinTr\cite{pointr}, which predicts the center points of missing regions based on the incomplete point cloud first and then fuses the incomplete point cloud with the predicted center points to generate complete point clouds, further advancing the field of point cloud completion. AdaPoinTr\cite{adapointr}, as an improved version of PoinTr, enhanced performance by developing an adaptive query generation mechanism and designing a novel denoising task during point cloud completion. However, despite explicitly predicting the center points of missing regions, these methods still rely solely on incomplete point clouds to predict the center points, lacking a global perspective. As a result, the reconstruction from incomplete point clouds to center points remains limited.\par
Given these limitations, we rethink the approach to point cloud completion and introduce a novel framework with corresponding modules to better address the challenges in point cloud completion tasks.

\subsection{Cascaded Networks}
Cascaded networks are a hierarchical framework widely used in computer vision\cite{cas_v_1, cas_v_2}. They consist of multiple stages, where each stage progressively generates the output. In point cloud processing, PCN\cite{pcn} adopts a two-stage coarse-to-fine structure to complete point cloud reconstruction. Many subsequent models\cite{cas_refine, pmp} also follow a similar multi-stage coarse-to-fine approach. These methods decompose the challenging task into multiple steps, leading to performance improvements. However, they all directly use features extracted from the coarse point cloud to assist in the subsequent refinement. And the information contained in the incomplete point cloud is inherently limited. Besides, when the goal is to refine the completion results from early stages, errors from the early stages may propagate to later stages, leading to error accumulation and ultimately degrading the final completion results. \par

\subsection{Knowledge Distillation}
Knowledge distillation is a method for model compression and acceleration, where a student model is trained to replicate the behavior of a pre-trained teacher model. Existing approaches fall into two categories based on the teacher’s knowledge source: enhanced model capacity\cite{dist1, dist2}, and additional data information\cite{dist4, dist5}. For the first category, a more complex teacher model provides richer knowledge but incurs significant computational and memory costs. For the second, additional data, which is often multimodal, serves as the knowledge source, yet acquiring such data poses practical challenges. Therefore, a key question we explore is whether it is possible to achieve effective knowledge distillation without introducing excessively complex models or relying on additional data sources. Moreover, when designing knowledge distillation strategies, it is crucial to consider the teacher-student paradigm gap\cite{dist_gap} and the potential shortcut\cite{shortcut} problem.

\section{METHOD}
\subsection{Network Architecture}
The goal of the point cloud completion task is to predict a complete point cloud from an incomplete one. Given an incomplete point cloud \( P_{\text{in}} = \{ p_i \}_{i=1}^{Q} \), where \( p_i \in \mathbb{R}^3 \) represents the 3D coordinates of a point, the task is to generate a complete point cloud \( P_{\text{out}} = \{ q_j \}_{j=1}^{M} \), where \( M \gg Q \). The overall framework of CasPoinTr is shown in Fig.\ref{fig:caspointr}. CasPoinTr adopts AdaPoinTr as the backbone and incorporates knowledge distillation and cascaded networks, which divides point cloud completion into two stages: Shape Reconstruction and Fused Completion. Let $\psi_1$, $\psi_2$ denote the models in Shape Reconstruction and Fused Completion, respectively. Let $\phi$ be the auxiliary feature encoder. The process of point cloud processing in CasPoinTr can be formulated as: \par
\begin{equation}
    y = \psi_2 \left(\phi \left( \psi_1(x)\right), x  \right) .
\end{equation} \par
In the following parts, we will provide a detailed description of these components.

\subsection{Auxiliary Completion}
To improve the point cloud completion process by providing additional information and minimizing error accumulation, we propose a cascaded networks structure based on an auxiliary completion approach. This design splits the completion task into two distinct stages: Shape Reconstruction and Fused Completion. The Shape Reconstruction stage focuses on recovering dense point cloud information, which serves as auxiliary data to guide the subsequent Fused Completion stage. By leveraging the results from Shape Reconstruction, Fused Completion gains access to richer information about the target, resulting in improved latent features and enhanced overall completion performance.\par
\begin{figure}[t]
\centering

\begin{subfigure}[b]{0.45\textwidth} 
   \includegraphics[width=\textwidth]{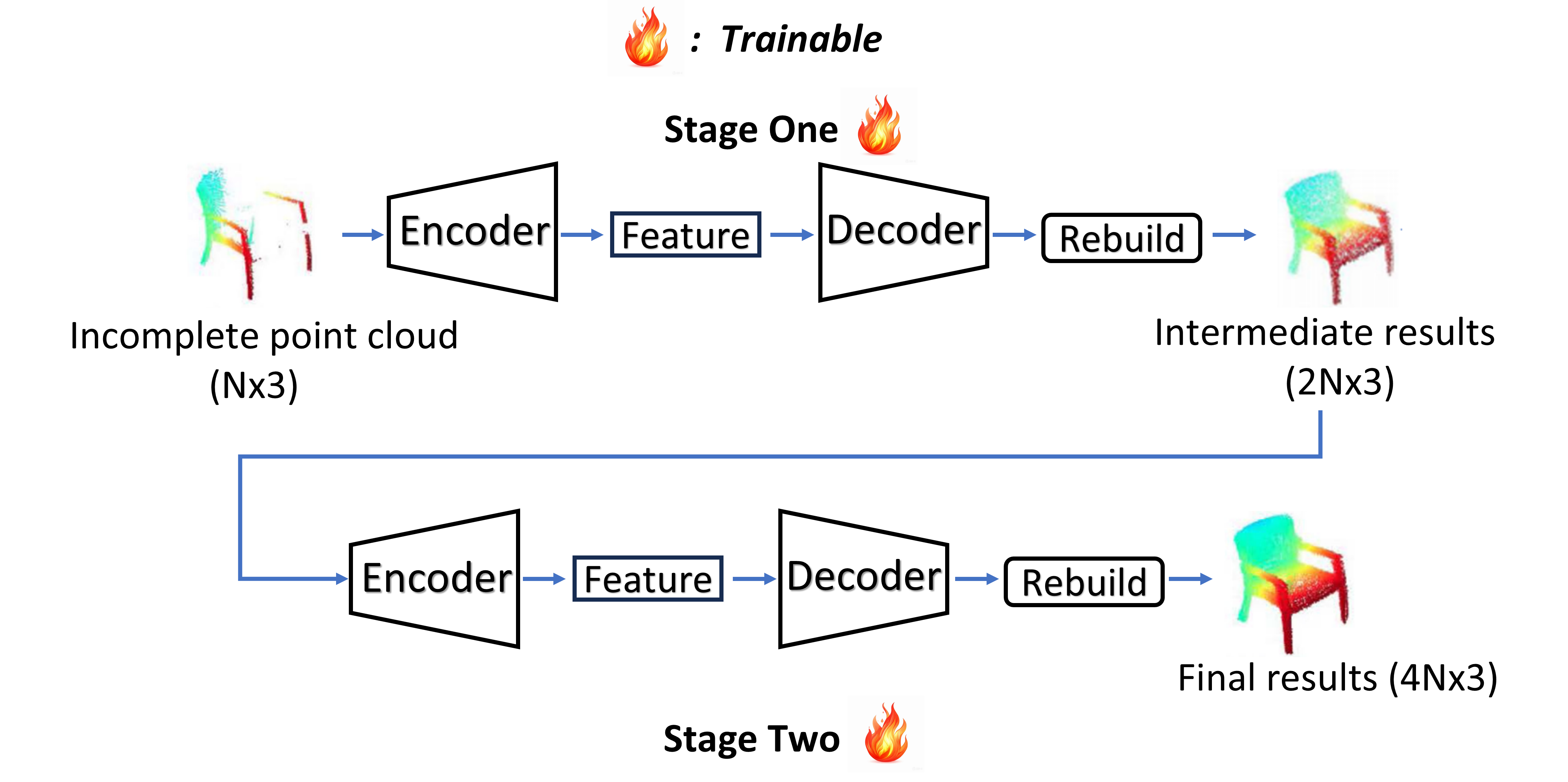}
   \caption{Simple Design: Progressive Completion.}
   \label{fig:sub1}
\end{subfigure}

\begin{subfigure}[b]{0.45\textwidth}
   \includegraphics[width=\textwidth]{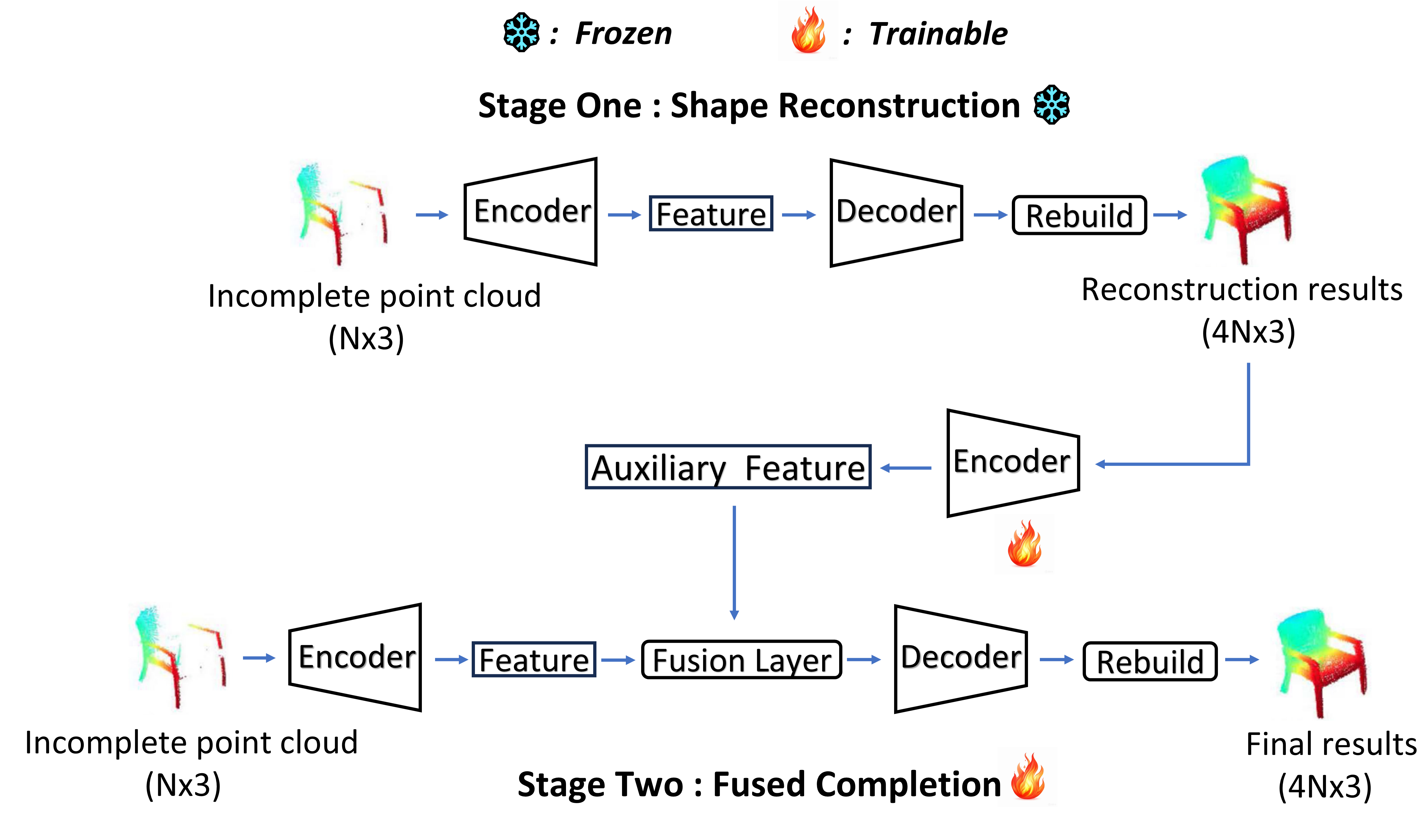}
   \caption{Our Proposal: Auxiliary Completion.}
   \label{fig:sub2}
\end{subfigure}

\caption{Different structure of cascaded networks. (a). Progressive Completion: The first stage performs 2× upsampling, and its results are used as input for the second stage, which also performs 2× upsampling.
(b). Auxiliary Completion: The completion results from the first stage provide auxiliary information to guide the second staga. The second stage still takes the incomplete point cloud as input.}
\label{fig:fig2}
\end{figure}

For tasks like 4× upsampling on ShapeNet-55, a naive approach would be to break the process into two consecutive 2× upsampling stages, which we term progressive completion (as depicted in Fig.\ref{fig:sub1}). While intuitive, this method suffers from error accumulation. To overcome this limitation, we propose an alternative strategy: in the Shape Reconstruction stage, we directly perform 4× upsampling to generate a dense point cloud that retains comprehensive shape details. In the Fused Completion stage, instead of using the reconstructed point cloud as direct input, we take the original incomplete point cloud as input and utilize the latent features extracted from the Shape Reconstruction results as auxiliary information. This approach, illustrated in Fig.\ref{fig:sub2}, is what we call auxiliary completion. Specifically, the encoder used for extracting the auxiliary features follows the architecture of AdaPointr’s encoder and is initialized with pretrained weights. The fusion layer performs channel-wise fusion by concatenating features along the channel dimension and processing them through a fully connected layer.\par
By executing direct 4× upsampling in Shape Reconstruction, we ensure the generated point cloud contains richer information. Meanwhile, in Fused Completion, relying on the Shape Reconstruction output solely as auxiliary data rather than as the primary point cloud input allows the model to benefit from enhanced latent features while mitigating the impacts of error accumulation. This cascaded design strikes an effective balance between providing additional information and maintaining output fidelity.\par

\subsection{The Teacher with Appropriate Privileged Inputs}
To enhance the student model's capability to estimate the overall shape from incomplete point clouds, we employ knowledge distillation with a teacher model that leverages carefully selected privileged inputs. This approach tackles two critical challenges: the teacher-student paradigm gap and the shortcut problem. The teacher model shares the same architecture as CasPoinTr and utilizes privileged inputs with the denser point cloud. The goal of knowledge distillation is to enable the student model to learn incomplete-complete associative knowledge from the teacher model in the training phase. \par
\begin{figure}[t]
\centering
\includegraphics[width=0.45\textwidth]{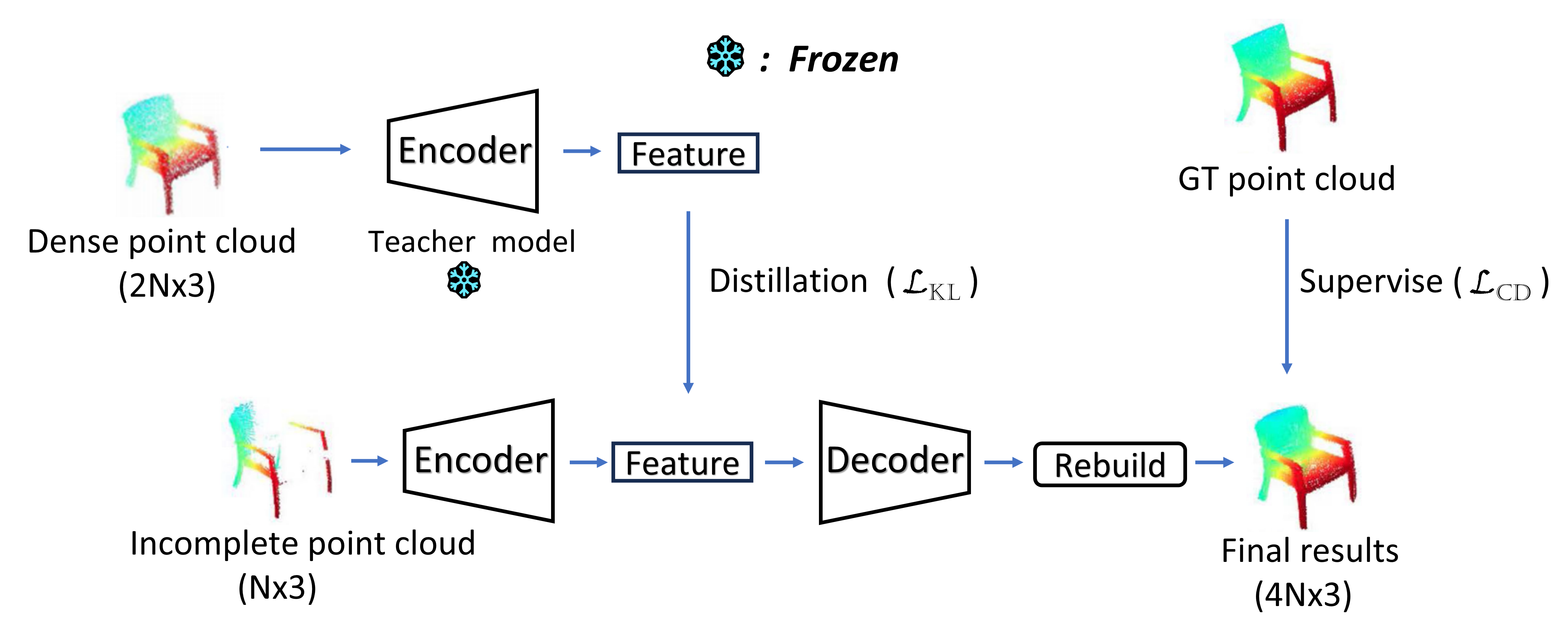}
\caption{Schematic diagram of knowledge distillation. The teacher model accepts the point cloud sampled from GT point cloud with resolutions of 2N×3 as input. The Kullback-Leibler divergence is used as loss function.}
\label{fig:dist}
\end{figure}

Regarding the resolution of privileged input point clouds, a straightforward approach is to use the same resolution as the Ground Truth (4N×3) as the privileged input. However, this would lead to shortcut issue\cite{shortcut} and an excessive gap\cite{dist_gap} in the teacher-student paradigm. To address these, we propose using the point cloud sampled from GT point cloud with a resolution of 2N×3 as privileged inputs for the teacher model. This intermediate resolution input provides adequate guidance for the student model—offering richer information than the low-resolution input (Nx3), while mitigating the paradigm gap and avoiding shortcut learning compared to the full-resolution input (4Nx3). In addition, for the auxiliary feature, we also employ knowledge distillation to reduce errors introduced by the reconstruction results. Here, the teacher model uses GT points as privileged inputs, with a resolution consistent with the reconstruction results. Regarding loss function, both the Kullback-Leibler divergence and the mean squared error have been applied in feature-level knowledge distillation for point clouds\cite{kl_dist}. In CasPoinTr, knowledge distillation is applied to the features output by the Encoder, and the Kullback-Leibler divergence is adopted as loss function. The schematic diagram of knowledge distillation is shown in Fig.\ref{fig:dist}. 

\begin{table*}[ht]
\centering
\caption{Point cloud completion results on ShapeNet-55, presenting Chamfer Distance (CD) under $\mathrm{CD}{-}\ell_1$ and $\mathrm{CD}{-}\ell_2$ (both scaled by 1000), F-Score@1\% for all 55 categories under Simple setting, along with detailed results for 8 specific categories.}
\label{tab:tab1}
\begin{tabular}{>{\centering\arraybackslash}p{2cm} >{\centering\arraybackslash}p{2cm} >{\centering\arraybackslash}p{2cm} | >{\centering\arraybackslash}p{2cm} >{\centering\arraybackslash}p{2cm} | >{\centering\arraybackslash}p{2cm} >{\centering\arraybackslash}p{2cm}}
\toprule
\multirow{2}{*}{Category} & \multicolumn{2}{c|}{$\mathrm{CD}{-}\ell_1$} & \multicolumn{2}{c|}{$\mathrm{CD}{-}\ell_2$} & \multicolumn{2}{c}{F-Score@1\%} \\
\cmidrule{2-7}
& AdaPoinTr & CasPoinTr & AdaPoinTr & CasPoinTr & AdaPoinTr & CasPoinTr \\
\midrule
Chair & 11.40 & \textbf{11.27} & 0.48 & \textbf{0.46} & 41.5\% & \textbf{42.4\%} \\
Airplane & 8.25 & \textbf{8.05} & 0.25 & \textbf{0.24} & 63.2\% & \textbf{64.0\%} \\
Car & 14.56 & \textbf{14.47} & 0.70 & \textbf{0.69} & 24.39\% & \textbf{24.84\%} \\
Sofa & 12.98 & \textbf{12.80} & 0.57 & \textbf{0.55} & 32.3\% & \textbf{32.8\%} \\
Skateboard & 8.51 & \textbf{8.39} & 0.25 & \textbf{0.23} & 58.8\% & \textbf{60.2\%} \\
Birdhouse & 15.95 & \textbf{15.08} & 1.00 & \textbf{0.79} & 27.4\% & \textbf{27.9\%} \\
Earphone & 13.42 & \textbf{12.57} & 0.65 & \textbf{0.63} & 37.2\% & \textbf{39.4\%} \\
Keyboard & 9.61 & \textbf{9.37} & 0.31 & \textbf{0.27} & 45.8\% & \textbf{47.2\%} \\
\midrule
Mean & 11.99 & \textbf{11.87} & 0.56 & \textbf{0.53} & 41.3\% & \textbf{41.4\%} \\
\bottomrule
\end{tabular}
\end{table*}

\begin{figure}[ht]
\centering

\begin{subfigure}[b]{0.45\textwidth} 
   \includegraphics[width=\textwidth]{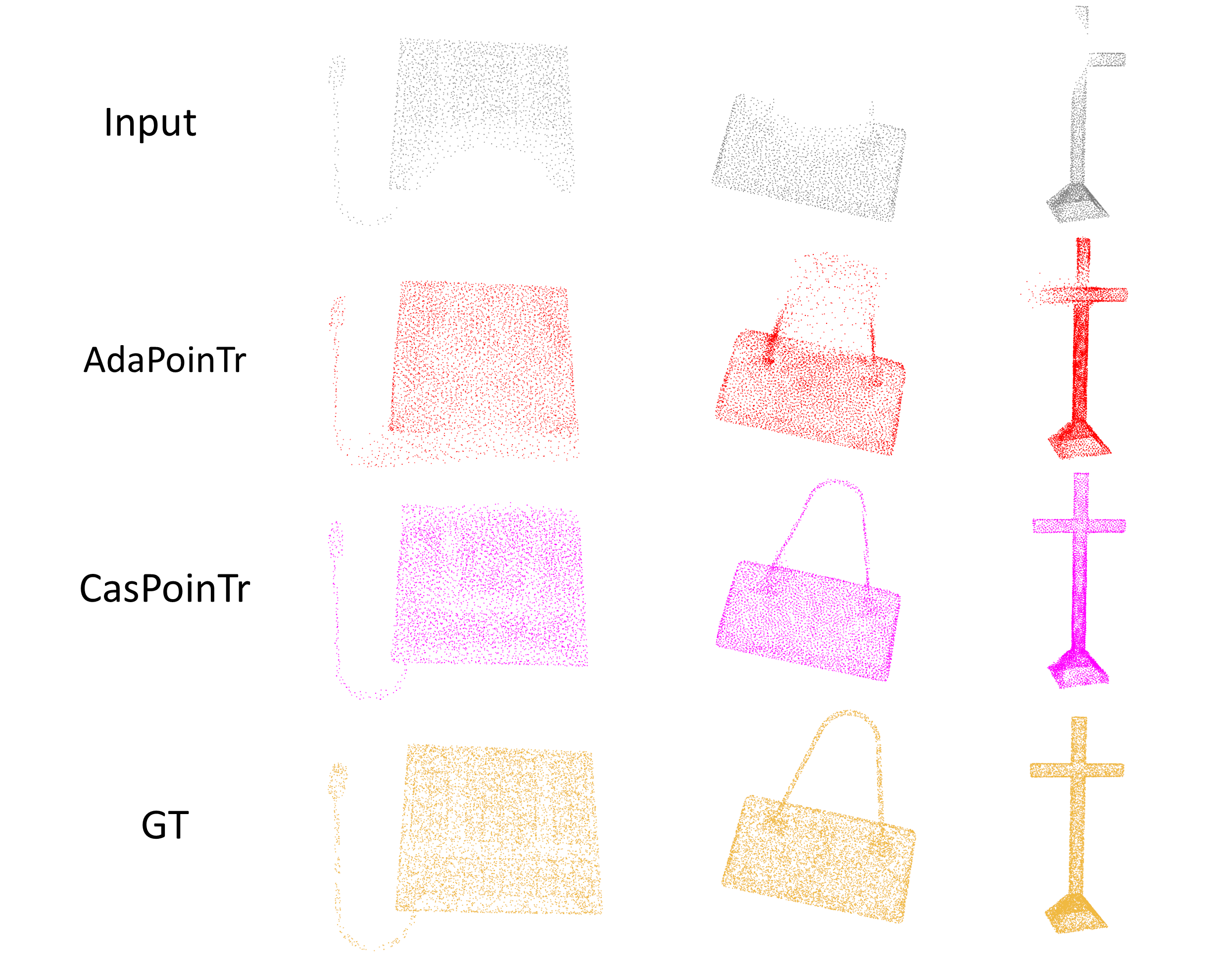}
   \caption{Visualization results on categories with a small number of samples. Categories from left to right are keyboard, bag, tower.}
   \label{fig:vis1}
\end{subfigure}

\begin{subfigure}[b]{0.45\textwidth}
   \includegraphics[width=\textwidth]{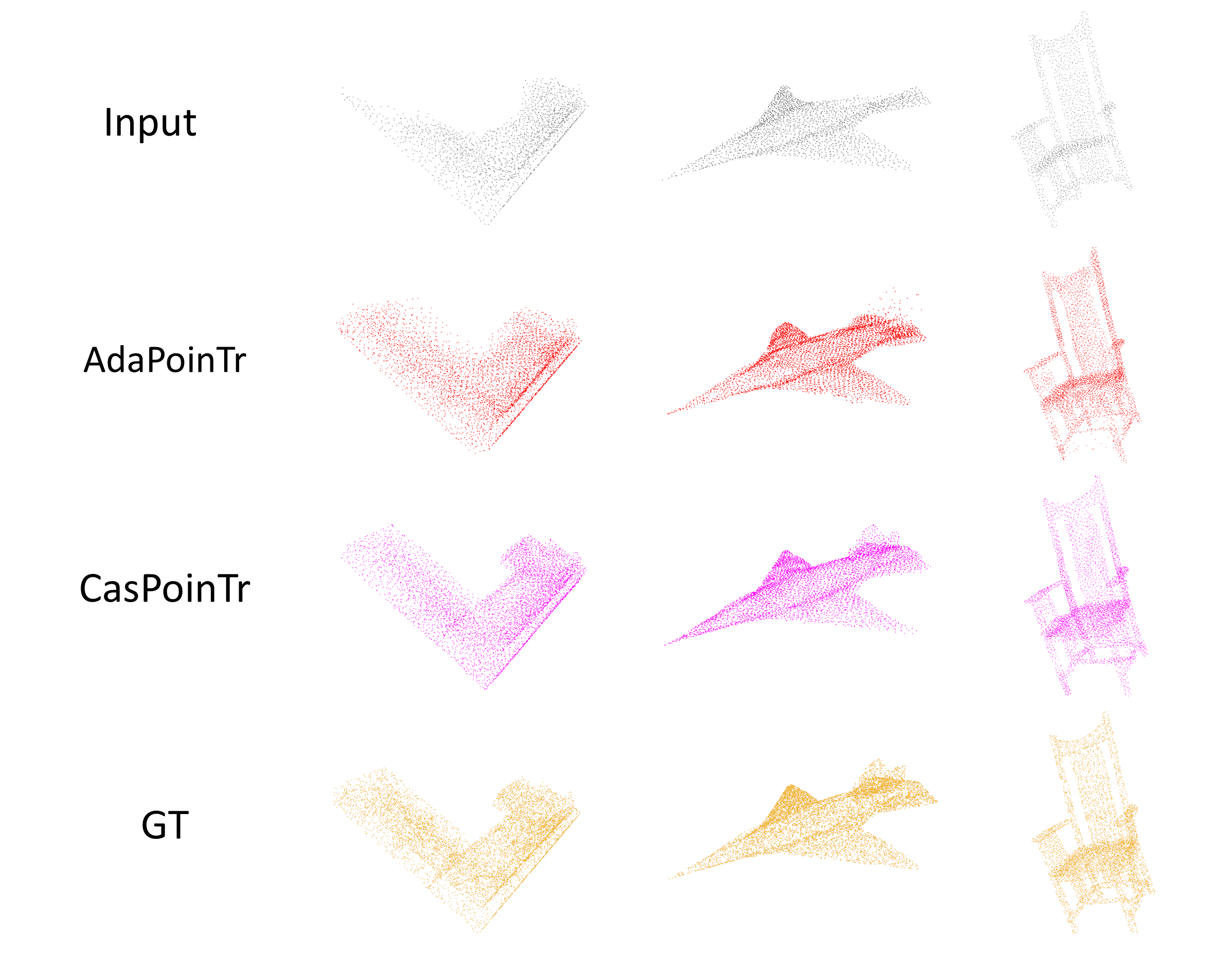}
   \caption{Visualization results  on categories with a large number of samples. Categories from left to right are sofa, airplane, chair.}
   \label{fig:vis2}
\end{subfigure}

\caption{Visualization of point cloud completion results on ShapeNet-55. The input data (first row) are obtained by first removing N points farthest from the viewpoint from 4N points. The input data is downsampled to N points as model's input, which is then processed using AdaPoinTr (second row) or CasPoinTr (third row), with the ground truth (GT) shown in the last row.}
\label{fig:visual}
\end{figure}


\subsection{Optimization}
We adopt the same loss functions as AdaPoinTr\cite{adapointr} and additionally introduce a distillation loss. Use $\mathcal{G}_s$ to represent the privileged input point cloud of Teacher B, which is sampled from the ground-truth complete point cloud. Let $\phi_A$ denote Teacher model A and $\phi_B$ denote Teacher model B. Then, we can write the teacher model's output as:\par

\begin{equation}
Z_A = \phi_A(\mathcal{G}), \quad  Z_B = \phi_B(\mathcal{G}_s).
\end{equation}\par

Let $Z_{Aux}$ denote the feature extracted from the Shape Reconstruction results, and let $Z_0$ denote the feature extracted by the encoder of the Fused Completion model. The distillation loss can be written as:\par
\begin{equation}
L_{KL_A}=D_{kl}(Z_A||Z_{Aux}), \quad L_{KL_B}=D_{kl}(Z_B||Z_{0}).
\end{equation}\par


Let $L_0$ be the objective funtion in AdaPoinTr. Our final loss function is the sum of the above terms.
\begin{equation}
L_{PC}=L_0+\lambda_1 L_{KL_A}+\lambda_2 L_{KL_B}.
\end{equation} \par

\section{EXPERIMENT}
To verify the effectiveness of our method, we conduct extensive experiments on ShapeNet-55\cite{shapenet} under simple and moderate settings.

\subsection{Dataset Details}
\textbf{ShapeNet-55:} ShapeNet-55 is a comprehensive benchmark designed to evaluate point cloud completion models across a diverse range of object categories. ShapeNet-55 includes objects from all 55 categories in the ShapeNet dataset. 
The dataset contains 41,952 models for training and 10,518 models for testing. For each object, 8,192 points are randomly sampled from the surface to generate the point cloud representation. Thus, ShapeNet-55 offers a robust and diverse platform for evaluating the performance of point cloud completion models.

\subsection{Implementation Details}
The architecture of the backbone follows the basic structure of AdaPoinTr. We adopt the Kullback-Leibler divergence\cite{kl} as the distillation loss function. Other loss functions are consistent with those used in AdaPoinTr. The evaluation metrics include chamfer distance (CD) and the F1 score, as described in \cite{metric}. We sample N points from the object as model's input. Specifically, to simulate real-world scenarios, we generate the input data by first eliminating the N points farthest from the viewpoint. Subsequently, we downsample the remaining point cloud to N points, which serves as the model's input. For fair comparison, we train AdaPoinTr and CasPoinTr using the same strategy and hyperparameter on V-100 GPUS. Teacher model A is pre-trained with a two-stage point cloud completion task using the model architecture shown in Fig.\ref{fig:sub2}, with training lasting for 300 epochs. Teacher model B is pre-trained with a single-stage point cloud completion task using the model architecture shown in Fig.\ref{fig:dist}. Notably, the input point cloud resolution is 2N×3, with training lasting for 600 epochs. Shape Reconstruction model use the checkpoints provided in \cite{adapointr}. Fused Completion model is trained for 300 epochs. N is set to 2048.

\begin{table}[t]
\centering
\caption{Model design analysis on ShapeNet-55 under Simple setting. We report $\mathrm{CD}{-}\ell_1$, $\mathrm{CD}{-}\ell_2$(both multiplied by 1000) and F-Score@1\%}
\begin{tabular}{    >{\centering\arraybackslash}p{2.5cm}  >{\centering\arraybackslash}p{1.75cm}  >{\centering\arraybackslash}p{0.75cm} >{\centering\arraybackslash}p{0.75cm} >{\centering\arraybackslash}p{0.75cm}}
\toprule
\multirow{2}{*}{Model}  & \multirow{2}{*}{Cascaded Design} & \multirow{2}{*}{$\mathrm{CD}{-}\ell_1$} & \multirow{2}{*}{$\mathrm{CD}{-}\ell_2$} & \multirow{2}{*}{F1-Score} \\
& & & & \\
\midrule
\multirow{2}{*}{AdaPoinTr} & \multirow{2}{*}{-} & \multirow{2}{*}{\textbf{11.99}} & \multirow{2}{*}{0.56} & \multirow{2}{*}{41.3\%}  \\ 
  &  &  &  &   \\

\hline \multirow{2}{*}{Cascaded Networks A} & \multirow{2}{*}{Progressive} & \multirow{2}{*}{15.45} & \multirow{2}{*}{0.88} & \multirow{2}{*}{35.5\%}  \\ 
& & & & \\

\hline \multirow{2}{*}{Cascaded Networks B} & \multirow{2}{*}{Auxiliary} & \multirow{2}{*}{12.02} & \multirow{2}{*}{\textbf{0.54}} & \multirow{2}{*}{\textbf{41.4\%}}  \\ 
& & & & \\

\bottomrule
\label{tab:tab2}
\end{tabular}
\end{table}

\begin{table}[t]
\centering
\caption{Model design analysis on ShapeNet-55 under Simple and Moderate settings. We report $\mathrm{CD}{-}\ell_1$, $\mathrm{CD}{-}\ell_2$(both multiplied by 1000 and averaged on two settings) and F-Score@1\%}
\begin{tabular}{    >{\centering\arraybackslash}p{2.75cm}  >{\centering\arraybackslash}p{1.5cm}  >{\centering\arraybackslash}p{0.75cm} >{\centering\arraybackslash}p{0.75cm} >{\centering\arraybackslash}p{0.75cm}}
\toprule
\multirow{2}{*}{Model}  & \multirow{2}{*}{Priority input} & \multirow{2}{*}{$\mathrm{CD}{-}\ell_1$} & \multirow{2}{*}{$\mathrm{CD}{-}\ell_2$} & \multirow{2}{*}{F1-Score} \\
& & & & \\
\midrule
\multirow{2}{*}{AdaPoinTr} & \multirow{2}{*}{-} & \multirow{2}{*}{12.61} & \multirow{2}{*}{\textbf{0.66}} & \multirow{2}{*}{40.3\%}  \\ 
  &  &  &  &   \\
\hline \multirow{4}{*}{Distill with MSE Loss} & \multirow{2}{*}{2Nx3} & \multirow{2}{*}{12.66} & \multirow{2}{*}{0.67} & \multirow{2}{*}{40.0\%}  \\ 
& & & & \\
  & \multirow{2}{*}{4Nx3} & \multirow{2}{*}{12.86} & \multirow{2}{*}{0.67} & \multirow{2}{*}{39.8\%}  \\
  & & & & \\
\hline \multirow{4}{*}{Distill with KL Loss} & \multirow{2}{*}{2Nx3} & \multirow{2}{*}{\textbf{12.55}} & \multirow{2}{*}{\textbf{0.66}} & \multirow{2}{*}{\textbf{40.4\%}}  \\ 
& & & & \\
  & \multirow{2}{*}{4Nx3} & \multirow{2}{*}{12.78} & \multirow{2}{*}{0.67} & \multirow{2}{*}{40.2\%}  \\
  & & & & \\

\bottomrule
\label{tab:tab3}
\end{tabular}
\end{table}

\begin{table}[t]
\centering
\caption{Ablation study on ShapeNet-55 under Simple setting. We report $\mathrm{CD}{-}\ell_1$, $\mathrm{CD}{-}\ell_2$(both multiplied by 1000) and F-Score@1\%}
\begin{tabular}{    >{\centering\arraybackslash}p{1.5cm}  >{\centering\arraybackslash}p{1.2cm}  >{\centering\arraybackslash}p{1.2cm}  >{\centering\arraybackslash}p{0.75cm} >{\centering\arraybackslash}p{0.75cm} >{\centering\arraybackslash}p{0.75cm}}
\toprule
\multirow{2}{*}{Model}  & \multirow{2}{*}{Cascaded}  & \multirow{2}{*}{Distillation} & \multirow{2}{*}{$\mathrm{CD}{-}\ell_1$} & \multirow{2}{*}{$\mathrm{CD}{-}\ell_2$} & \multirow{2}{*}{F1-Score}  \\
& & & & \\
\midrule
\multirow{2}{*}{A} & \multirow{2}{*}{-} & \multirow{2}{*}{-} & \multirow{2}{*}{11.99} & \multirow{2}{*}{0.56} & \multirow{2}{*}{41.3\%}  \\ 
  &  &  &  & &   \\
\hline \multirow{2}{*}{B}  & \multirow{2}{*}{\checkmark} & \multirow{2}{*}{-} & \multirow{2}{*}{12.02} & \multirow{2}{*}{0.54} & \multirow{2}{*}{\textbf{41.4\%}}  \\ 
& & & & & \\

\hline \multirow{2}{*}{CasPoinTr}  & \multirow{2}{*}{\checkmark} & \multirow{2}{*}{\checkmark} & \multirow{2}{*}{\textbf{11.87}} & \multirow{2}{*}{\textbf{0.53}} & \multirow{2}{*}{\textbf{41.4\%}}  \\ 
& & & & & \\

\bottomrule
\label{tab:tab4}
\end{tabular}
\end{table}

\subsection{Main Results}
We compare our proposed approach with the state-of-the-art method AdaPoinTr\cite{adapointr}. The experimental results are shown in Tab.\ref{tab:tab1}. We report Chamfer Distance under $\mathrm{CD}{-}\ell_1$, $\mathrm{CD}{-}\ell_2$ and F1-Score for all 55 categories under Simple setting, along with detailed results for 8 specific categories. It can be observed that CasPoinTr achieves superior performance compared to AdaPoinTr, with a $\mathrm{CD}{-}\ell_1$ of 11.87, a $\mathrm{CD}{-}\ell_2$ of 0.53, and an F1 Score of 41.4\% on ShapeNet-55. These results highlight the effectiveness of CasPoinTr even in a diverse situation. \par
What's more, to investigate the effectiveness of different designs for knowledge distillation and cascaded networks, we conduct exhaustive model design analysis on these key components of CasPoinTr.\par
We conducted experiments using the architectures shown in Fig.\ref{fig:sub1} and Fig.\ref{fig:sub2} as cascaded networks structures, and the results are presented in Tab.\ref{tab:tab2}. It can be observed that the architecture in Fig.\ref{fig:sub1} suffers from performance degradation due to error accumulation. In contrast, the architecture in Fig.\ref{fig:sub2} effectively mitigates the issue of error accumulation. With the assistance of auxiliary features, the model achieves significant performance improvement.\par
We explored the design of knowledge distillation using the single-stage architecture shown in Fig.\ref{fig:dist}. Specifically, we begin by randomly removing between N and 2N points that are farthest from the viewpoint. Then we downsample the remaining point cloud to N points as model's input. Experimental results for the teacher model with privileged inputs of different resolutions and using either the Kullback-Leibler divergence or the mean squared error as loss function are presented in Tab.\ref{tab:tab3}. It can be observed that using point clouds with a resolution of 2N×3 as privileged inputs better enhances the model's performance. This is attributed to its effectiveness in avoiding shortcuts and maintaining a smaller teacher-student paradigm gap. Employing the Kullback-Leibler divergence as loss function effectively improves the model's performance, whereas using the mean squared error as loss function degrades the model's performance. \par
The visualization results are shown in Fig.\ref{fig:visual}. We report three categories with a small number of samples in Fig.\ref{fig:vis1} and three categories with a large number of samples in Fig.\ref{fig:vis2}. It can be observed that CasPoinTr performs better in both shape recovery and fine-grained detail preservation.\par

\subsection{Ablation Study}
The design analysis results of cascaded networks and knowledge distillation are presented in Tab.\ref{tab:tab2} and Tab.\ref{tab:tab3}. To further examine their effectiveness, we conduct a detailed ablation study, with the results summarized in Tab.\ref{tab:tab4}. Model A adopt the same architecture as AdaPoinTr without cascaded networks or knowledge distillation design. We introduce cascaded networks (model B) and we see an improvement in $\mathrm{CD}{-}\ell_2$ and F1-Score. When both using cascaded netwrok and knowledge distillation (CasPoinTr), we observe an further  improvement, which clearly demonstrates the effectiveness of cascaded networks and knowledge distillation.

\section{CONCLUSION}
In this paper, we introduce CasPoinTr, a novel cascaded networks framework for point cloud completion that seamlessly integrates Shape Reconstruction and Fused Completion to enhance the completion quality. By employing a cascaded structure, CasPoinTr generates intermediate representations to guide Fused Completion, ensuring robust and accurate reconstruction of missing regions. Additionally, we incorporate knowledge distillation, allowing a teacher model with privileged access to high-resolution point clouds to guide the student model during training, improving its ability to estimate the overall shape and infer missing regions more accurately. The proposed method can be transferred and applied to various state-of-the-art backbones, exhibiting excellent generalizability and scalability.\par
Our experimental results on ShapeNet-55 demonstrate the superiority of CasPoinTr over existing state-of-the-art methods. Compared to previous approaches, CasPoinTr achieves better performance in terms of Chamfer Distance and F1-Score, demonstrating a significant improvement in both shape reconstruction and fine-grained detail preservation. \par
While CasPoinTr demonstrates promising results, there remains room for improvement. Future research could focus on optimizing the cascaded networks to strike a better balance between computational efficiency and accuracy. Such enhancements would further increase its practical utility in real-world applications.\par





\end{document}